%% file: ___preprint___newWhen_Harnesses_Lose_the_Signal__Causal_Information_Diagnostics_for_LLM_Agents.tex
\documentclass{article} 
\usepackage{iclr2027_conference,times}

\input{math_commands.tex}

\usepackage{graphicx}
\usepackage{url}
\usepackage{amsmath}
\usepackage{amssymb}
\usepackage{booktabs}
\usepackage{placeins}
\definecolor{citecolor}{HTML}{0071bc}
\usepackage[pagebackref=false,breaklinks=true,letterpaper=true,colorlinks,citecolor=citecolor,bookmarks=false]{hyperref}

\title{When Harnesses Lose the Signal: Causal Evaluation of Recovery in LLM Agents}

\author{
Shuyao Xiao$^{1,2}$, Shengling Wang$^{1}$, Xuan Chen$^{2}$,
Ke Chao$^{1}$, Ming Cui$^{2}$, Feifei Qian$^{1}$,\\
Chaoyang Mei$^{2}$, Fanlin Meng$^{2}$,
Ziming Yu$^{1}$, Junxi Yin$^{2}$ \\
$^{1}$School of Artificial Intelligence, Beijing Normal University \quad
$^{2}$Ke Holdings \\
\texttt{xiaoshuyao@mail.bnu.edu.cn}
}

\iclrfinalcopy 
\begin{document}

\maketitle
\lhead{Preprint}
\begin{abstract}
Large language model agents rely on external harnesses to pass information between the model and its environment and to recover from execution errors. Yet recovery is usually judged only by average task success. This hides an important tension. The same operation can rescue a failing trajectory or disrupt one that would otherwise succeed. We frame recovery as a causal decision problem. Starting from the same execution state, we compare what happens with and without recovery, separate rescue from harm, and study how the value of recovery changes over time. We then introduce the Causal Intervention Router (CIR), a lightweight policy that uses information available before recovery to decide when intervention is worthwhile. On long-horizon ALFWorld tasks with Qwen3-14B, CIR raises success from 70.33\% to 73.33\%, a gain of \(3.00\) percentage points. It leaves all evaluated trajectories with correct observations untouched. Additional controls show that the benefit of recovery cannot be explained solely by the new observation returned by the environment. These results provide a practical way to evaluate recovery and apply it selectively. 
\end{abstract}

\section{Introduction}
\label{sec:introduction}

Large language model agents do not act alone. An external harness passes observations between the model and its environment, maintains context, and coordinates long sequences of actions~\citep{packer2023memgpt,wu2024autogen,yang2024sweagent,yao2026harnessbenchmeasuringharnesseffects}. If an observation is outdated or missing, the agent may act on the wrong information, and the error can spread through later steps~\citep{yao2023react,shah2026causalagentreplay,ma2026longhorizonharnessadvancinglonghorizonagents}. A harness can respond by querying the environment again, which is called \emph{refresh}.

Refresh is not always helpful. It may turn a failure into a success, but it may also derail a trajectory that would otherwise finish the task. We call these outcomes \emph{rescue} and \emph{harm}. Existing methods improve recovery through reflection, external feedback, or search~\citep{shinn2023reflexion,madaan2023selfrefine,gou2024critic,zhou2024lats}. Evaluations, however, usually report an overall success rate or identify where failures occur~\citep{ma2024agentboard,cemri2025mast,zhang2025which}. These summaries do not show which trajectories recovery changes. The same average gain can hide very different numbers of rescues and harms.

This creates two challenges. First, an ordinary run shows only one outcome. The agent either recovers or continues without recovery~\citep{holland1986statistics}. Without the alternative outcome, we cannot tell whether recovery caused the final result. Second, timing matters. After an error, later actions and feedback change the situation. An intervention that helps now may be ineffective or harmful later. A useful evaluation must compare both choices from the same state and repeat that comparison at different times.

Together, these challenges expose a gap between evaluating a recovery operation and deciding when to use it. A positive average effect does not imply that recovery is safe to apply broadly. The same average can result from a few reliable rescues or from many rescues offset by many harms. A modest average can also hide a large benefit in a recognizable subset of states. The central question is whether refresh works for a given trajectory and when its expected benefit outweighs its risk.

\begin{figure}[t]
    \centering
    \includegraphics[width=\linewidth]{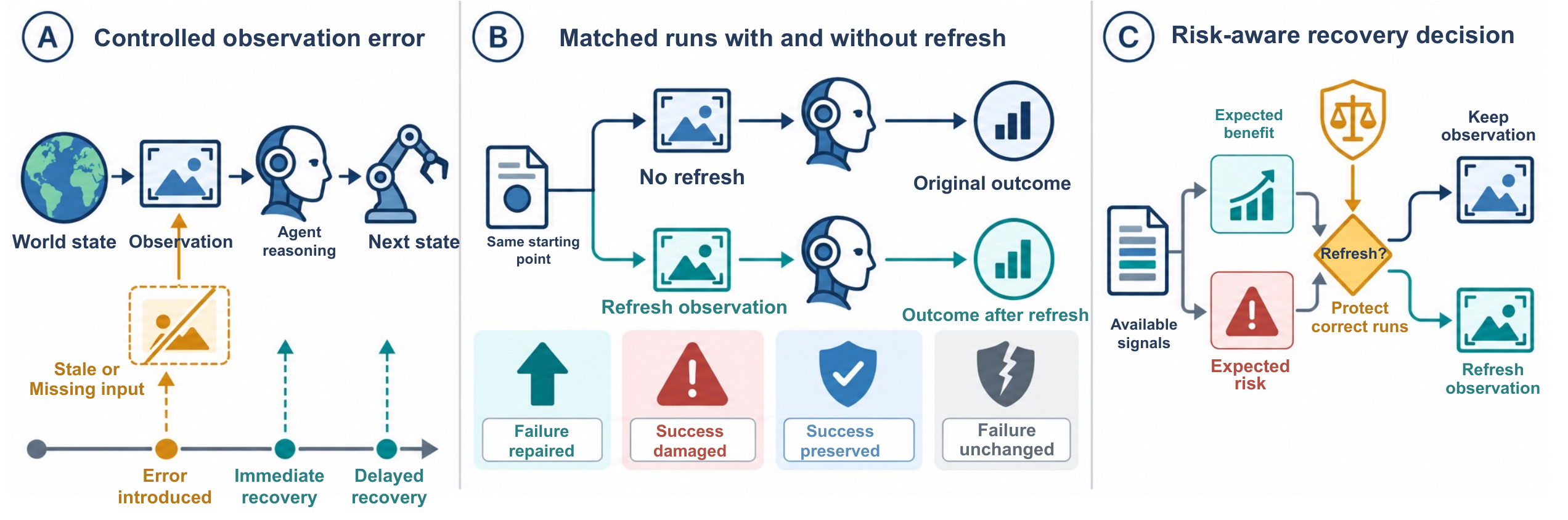}
    \caption{
    Overview of our framework.
    (A) We introduce controlled observation errors and try recovery at
    different times. (B) From the same execution point, paired runs with
    and without refresh reveal rescue, harm, and unchanged outcomes.
    (C) CIR uses information available before intervention to decide
    whether the expected benefit of refresh outweighs its risk.
    }
    \label{fig:overview}
\end{figure}

Figure~\ref{fig:overview} shows our approach. We first introduce controlled observation errors. From the same execution point, we create two continuations, one with refresh and one without. This paired comparison reveals whether refresh rescues the task, harms it, or leaves the outcome unchanged. We also compare refresh variants to identify what drives the effect. We then use these paired outcomes to train the Causal Intervention Router (CIR). CIR decides whether to refresh using information available at decision time. It also limits harm to trajectories with correct observations.

The paired design serves two purposes. As an evaluation tool, it measures the effect of recovery while holding the earlier trajectory fixed. As a source of supervision, it reveals the states in which recovery changes the final outcome. The same paired outcomes used for evaluation also train CIR. At test time, it uses only information available in a live run.

Experiments on long-horizon ALFWorld tasks show that the value of refresh varies across errors, intervention times, and trajectories. They also show that a newly returned observation is not the sole source of the benefit. CIR turns this evidence into a selective policy. For Qwen3-14B, it improves success from 70.33\% to 73.33\% and does not refresh any evaluated clean trajectory. Our main contributions are as follows.
\begin{enumerate}
    \item We introduce a paired framework for evaluating harness recovery. By continuing from the same pre-recovery state with and without an intervention, the framework attributes changes in the final outcome to recovery and separates rescue from harm. Comparisons across times and recovery variants show when refresh helps and which parts of the operation matter.

    \item We introduce CIR, which uses the paired evidence to decide when recovery is worth the risk. It requires no retraining of the underlying agent and uses only information available before intervention. On Qwen3-14B, CIR improves overall success by \(3.00\) percentage points. The gain reaches \(9.33\) points when the agent receives a more severely outdated observation. All evaluated clean trajectories remain untouched.
\end{enumerate}

\section{Related Work}

\paragraph{Agent Harnesses and Recovery Mechanisms.}
Agent harnesses organize interactions among models, tools, and environments, so their design can directly affect task performance~\citep{yao2026harnessbenchmeasuringharnesseffects}. AutoGen~\citep{wu2024autogen} provides composable conversation patterns, SWE-agent~\citep{yang2024sweagent} studies agent and computer interfaces, and StateFlow~\citep{wu2024stateflow} uses explicit states to manage long tasks. Recovery methods revise execution through reflection, feedback, or search~\citep{li2026reseek}. Reflexion~\citep{shinn2023reflexion} learns from task feedback, Self-Refine~\citep{madaan2023selfrefine} revises outputs iteratively, CRITIC~\citep{gou2024critic} verifies outputs with external tools, and LATS~\citep{zhou2024lats} combines feedback with tree search. These methods improve how agents recover, but their aggregate evaluations do not identify the trajectories changed by recovery or the successful trajectories that recovery disrupts. This gap matters because self-correction without reliable feedback can degrade an initially correct response~\citep{huang2024selfcorrect}. We instead measure rescue and harm from the same starting state and use those outcomes to learn when recovery should be applied.

\paragraph{Agent Failure Diagnosis and Causal Intervention.}
Process-level evaluation studies where and why agent failures occur~\citep{barke2026agentrxdiagnosingaiagent}. AgentBoard~\citep{ma2024agentboard} tracks fine-grained progress, MAST~\citep{cemri2025mast} provides a failure taxonomy for multi-agent systems, Who\&When~\citep{zhang2025which} identifies responsible agents and steps, and CatchBench~\citep{zhao2026catchbench} audits failures from configurations, live prefixes, and completed traces.
Counterfactual approaches move from describing failures toward tracing or repairing them. In memory-augmented dialogue, \citet{xiao2026errorsmemoriescausalpathway} use four counterfactual trajectories to
separate error propagation through memory updates and subsequent questions. Causal Agent Replay~\citep{shah2026causalagentreplay} estimates the contribution of individual steps, CausalFlow~\citep{bonagiri2026causalflow} generates targeted repairs, HarnessFix~\citep{chen2026harnessfix} diagnoses and patches harness flaws, and DoVer~\citep{ma2025dover} tests debugging hypotheses through intervention. These methods localize causes or repair observed failures. They do not directly answer our question. From the same current state, will a given recovery operation improve or damage the final outcome? When should it be applied? We answer these questions with paired runs and a policy that limits observed harm on clean trajectories.

This distinction separates our work from both failure localization and repair generation. We keep the recovery operation fixed and study its causal value across states and times. This makes rescue and harm directly comparable and allows the same evidence used for evaluation to train a deployment-time decision rule.

\section{Paired Counterfactual Evaluation of Recovery}
\label{sec:counterfactual_evaluation}

An aggregate success rate cannot tell us whether recovery changes a particular trajectory. We therefore compare recovery with continued execution from the same task state. Controlled changes to the observation make the error explicit, while paired runs isolate the recovery decision and reveal both rescue and harm (Figure~\ref{fig:overview}(A--B)).

\subsection{Perturbation and Recovery Setting}
\label{sec:intervention_setting}

The unit of evaluation is a task instance, denoted by \(i\). At a predefined point in the run, the harness keeps the environment unchanged and modifies only the observation shown to the agent. The observation condition \(e\) is clean, stale, or missing. A clean observation preserves the current information. A stale observation replaces it with information from an earlier state, and a missing observation removes its content. In this way, we change the information available to the agent without changing the underlying task.

The recovery delay \(d\) is the number of environment actions between the modified observation and the recovery operation \(m\). A refresh queries the environment again and gives the agent the current observation. Replanning instead asks the agent to reconsider its next actions using information already in context. The triple \((e,d,m)\) therefore specifies what information the agent receives, when recovery occurs, and which recovery operation is used. Section~\ref{sec:exp_setup} gives the concrete settings.

\subsection{Counterfactual Comparison}
\label{sec:counterfactual_comparison}

For each task \(i\) and observation condition \(e\), execution continues until delay \(d\). We then branch the run from exactly the same state. One branch continues without recovery. The other applies operation $m$. The branches share the environment state and the full execution history before this decision.

Because the two branches use the same state, history, model, and decoding settings, their difference is attributable to the recovery decision for the states studied here.

Using potential-outcome notation~\citep{rubin1974estimating}, let $Y_i^0(e)$ and $Y_i^m(e,d)$ denote the binary task outcomes without recovery and with operation $m$ at delay $d$, respectively. Under deterministic replay, the no-recovery continuation is shared across delay comparisons, so we suppress $d$ in $Y_i^0(e)$. The instance-level effect of recovery is
\begin{equation}
\tau_i^m(e,d)
=
Y_i^m(e,d)-Y_i^0(e).
\label{eq:instance_recovery_effect}
\end{equation}

\subsection{Rescue and Harm}
\label{sec:rescue_and_harm}

We distinguish two directions of recovery effects. A \emph{rescue} occurs when recovery turns a failure into a success, whereas \emph{harm} occurs when recovery turns a success into a failure.

Let \(\mathbf{1}[\cdot]\) denote the indicator function, which equals 1 when the enclosed condition holds and 0 otherwise. The rescue indicator \(\mathcal{R}_i^m(e,d)\) and harm indicator \(\mathcal{H}_i^m(e,d)\) for instance \(i\) are defined as
\begin{equation}
\mathcal{R}_i^m(e,d)
=
\mathbf{1}\!\left[
Y_i^0(e)=0,\;
Y_i^m(e,d)=1
\right],
\mathcal{H}_i^m(e,d)
=
\mathbf{1}\!\left[
Y_i^0(e)=1,\;
Y_i^m(e,d)=0
\right].
\label{eq:rescue_harm}
\end{equation}
If both continuations succeed, recovery preserves success. If both fail, recovery leaves the failure unchanged. Together with rescue and harm, these cases form the four possible outcomes of a paired comparison. The instance-level recovery effect can therefore be written as
\begin{equation}
\tau_i^m(e,d)
=
\mathcal{R}_i^m(e,d)
-
\mathcal{H}_i^m(e,d).
\label{eq:recovery_decomposition}
\end{equation}
The average recovery effect equals the rescue rate minus the harm rate. The same net effect can arise from rare changes or from frequent changes in both directions. The average alone cannot distinguish these cases. The paired outcomes also provide supervision for learning when recovery is more likely to rescue than harm a trajectory.

This view changes what counts as a successful recovery method. A method should not be judged only by how often it succeeds after intervention, because many of those trajectories may already have succeeded without intervention. Instead, evaluation should ask whether the method changes the outcome, in which direction, and at what point in the trajectory. The paired design answers all three questions with the same set of outcomes.

\section{From Evaluation to Selective Recovery}
\label{sec:selective_recovery}

Paired evaluation tells us whether recovery helped only after both branches have finished. In a live run, the harness must decide before either outcome is known. CIR estimates the value of recovery from the current state. It refreshes only when the expected benefit justifies the risk. A refresh obtains a new observation from the environment and adds it to the context. CIR refreshes at most once per trajectory (Figure~\ref{fig:overview}(C)).

CIR makes a simple distinction. Detecting that something looks wrong is different from predicting that a particular response will help. An anomaly detector addresses the first question. A recovery policy must also answer the second. Some erroneous trajectories are not rescued by refresh, and some apparently normal trajectories can be harmed by it. CIR combines error detection with separate predictions of task success under refresh and continued execution.

\subsection{Decision Objective}
\label{sec:decision_objective}

Let $X_{i,d}$ denote the information available before recovery at delay $d$. The decision $\pi(X_{i,d})\in\{0,1\}$ indicates whether to refresh or continue for now. Let $Y_i^\pi(e)$ denote the final binary outcome under this sequential policy. If no refresh is triggered, $Y_i^\pi(e)=Y_i^0(e)$. Otherwise, it equals the outcome after refresh at the policy-selected delay.

Let \(\epsilon\in[0,1]\) denote the allowed harm rate on clean trajectories, and let \(\mathbb{E}[\cdot]\) denote expectation over task instances under the specified observation condition. We denote an optimal policy by \(\pi^\star\) and formulate selective recovery as
\begin{equation}
\begin{aligned}
\pi^\star
&=
\arg\max_{\pi}
\mathbb{E}\!\left[
Y_i^\pi(e)-Y_i^0(e)
\mid e\neq\mathrm{clean}
\right],\\
\text{subject to}\qquad
&
\mathbb{E}\!\left[
Y_i^0(e)\bigl(1-Y_i^\pi(e)\bigr)
\mid e=\mathrm{clean}
\right]
\leq \epsilon .
\end{aligned}
\label{eq:selective_recovery_objective}
\end{equation}
The objective rewards improvement when the observation is wrong, relative to never recovering. Maximizing gain alone could encourage frequent refreshes that disrupt correct executions. The second line limits this risk. Its value is the fraction of clean trajectories that succeed without recovery but fail under the policy. At runtime, neither possible outcome is known. CIR must therefore estimate the value of refresh from information available before the decision.

\subsection{Estimating Recovery Value}
\label{sec:estimating_recovery_value}

Detecting a suspicious observation is not enough. An error may exist even when refresh would not improve the final outcome. CIR separates two questions. Did the agent receive a faulty observation? How likely is the task to succeed with or without refresh? It answers them using execution behavior, consistency between actions and observations, task progress, and contextual information.

Let \(x\) denote a value of \(X_{i,d}\), and let \(\Pr(\cdot\mid\cdot)\) denote conditional probability. We use \(p_{\mathrm{err}}(x)\) for the probability that the trajectory has received an erroneous observation, \(p_0(x)\) for the probability of success without subsequent recovery, and \(p_1(x)\) for the probability of success after immediate refresh. Denoting refresh by \(\mathrm{ref}\) and its final outcome at delay \(d\) by \(Y_i^{\mathrm{ref}}(e,d)\), we define
\begin{equation}
\begin{aligned}
p_{\mathrm{err}}(x)
&=
\Pr\!\left(e\neq\mathrm{clean}\mid X_{i,d}=x\right),\\
p_0(x)
&=
\Pr\!\left(Y_i^0(e)=1\mid X_{i,d}=x\right),\\
p_1(x)
&=
\Pr\!\left(Y_i^{\mathrm{ref}}(e,d)=1\mid X_{i,d}=x\right).
\end{aligned}
\label{eq:cir_outcome_probabilities}
\end{equation}
CIR estimates these quantities with three models supervised by the observation condition, the no-recovery outcome, and the refresh outcome in the paired data from Section~\ref{sec:counterfactual_evaluation}. The two outcome models follow a T-learner-style design and estimate the two decision outcomes separately~\citep{kunzel2019metalearners}. This separation prevents high anomaly probability from being treated as evidence that refresh is beneficial. For brevity, we use the same notation for the probabilities and their estimates.

To account for potential rescue and harm, we construct a rescue score $s_{\mathrm R}(x)$ and a harm score $s_{\mathrm H}(x)$.
\begin{equation}
s_{\mathrm R}(x)
=
\bigl(1-p_0(x)\bigr)p_1(x),
\qquad
s_{\mathrm H}(x)
=
p_0(x)\bigl(1-p_1(x)\bigr).
\label{eq:cir_rescue_harm_scores}
\end{equation}
The rescue score is high when continued execution is likely to fail and refresh is likely to succeed. The harm score is high when continued execution is likely to succeed and refresh is likely to fail. These scores help identify states in which refresh is more promising. The marginal success probabilities do not determine the probability that the same task is rescued or harmed~\citep{tian2000probabilities}. We therefore do not interpret these scores as rescue or harm probabilities.

Let $\lambda>0$ denote the harm penalty. Let $u_\lambda(x)$ denote the utility obtained by subtracting the weighted harm score from the rescue score.
\begin{equation}
u_\lambda(x)
=
s_{\mathrm R}(x)-\lambda s_{\mathrm H}(x).
\label{eq:cir_utility}
\end{equation}
When \(\lambda=1\), this expression reduces to \(p_1(x)-p_0(x)\), the difference in predicted success probabilities between refresh and no recovery. Increasing \(\lambda\) lowers the utility of states with high harm scores, making refresh less likely to be triggered in those states. Thus, \(\lambda\) controls the weight assigned to harm in each decision, whereas \(\epsilon\) limits the fraction of all clean trajectories that the policy turns from success into failure.

\subsection{Dynamic Recovery Decisions}
\label{sec:dynamic_recovery_decisions}

The value of recovery can change as the agent takes more actions and receives more feedback. CIR therefore reconsiders refresh at several candidate times instead of committing to one fixed delay. At each time, it compares an immediate refresh with continuing without later recovery. If it chooses to continue, it can reconsider after the state has changed.

Let $p_{\mathrm{fail}}(x)=1-p_0(x)$ denote the predicted probability of failure without subsequent recovery. Let $\alpha$, $\beta$, and $\gamma$ denote the thresholds for the error probability, failure probability, and utility score. Let $d_{\min}$ denote the earliest allowed refresh delay. The indicator function $\mathbf{1}[\cdot]$ was defined in Section~\ref{sec:counterfactual_evaluation}. The decision rule before any refresh has occurred is
\begin{equation}
\pi(X_{i,d})
=
\mathbf{1}\!\left[
\begin{aligned}
&d\geq d_{\min},\\
&p_{\mathrm{err}}(X_{i,d})\geq\alpha,\\
&p_{\mathrm{fail}}(X_{i,d})\geq\beta,\\
&u_\lambda(X_{i,d})\geq\gamma
\end{aligned}
\right].
\label{eq:cir_decision_rule}
\end{equation}
CIR checks candidate times in chronological order and refreshes at the first time when all conditions are satisfied. No further recovery is triggered afterward. If no candidate time satisfies the conditions, the task proceeds without recovery.

We select the decision thresholds on development data. Among the settings whose observed harm rate on clean trajectories does not exceed \(\epsilon\), we choose the one with the largest net gain under erroneous observations. We then apply that fixed policy to new tasks. This procedure is related to empirical welfare maximization~\citep{kitagawa2018treated}, with paired recovery outcomes defining the benefit and the clean-trajectory results defining the risk limit.

\section{Experiments}
\label{sec:experiments}

Our experiments answer three questions. When does refresh help or harm? Which parts of refresh produce the benefit? Can CIR use this evidence to recover selectively on new tasks? We first describe the common setup and then answer each question.

\subsection{Experimental Setup}
\label{sec:exp_setup}

\paragraph{Task, agent, and perturbations.} We use ALFWorld~\citep{shridhar2021alfworld}, where an agent completes long household tasks through text actions and observations. Our agent is Qwen3-14B~\citep{yang2025qwen3}, used with a fixed ReAct-style harness~\citep{yao2023react} and greedy decoding. Each run is limited to 49 model turns and 64 environment actions. We generated the agent trajectories on eight NVIDIA H200 GPUs. We first run each task without an injected error or recovery. We call this the \emph{factual trajectory}. The observation is modified after 25\% of the environment actions in that trajectory.

The primary perturbation is a two-step stale observation, which replaces the current observation with the observation from two environment actions earlier. Additional analyses use a one-step stale observation from one environment action earlier and a missing observation whose content is replaced by a fixed unavailable-observation message. A refresh executes an additional \texttt{look} action and appends the returned current observation to the transcript. We test recovery delays \(d\in\{0,1,2,4\}\).

\paragraph{Paired runs and data partitions.} Each comparison starts two runs from the same environment state and model input, with identical model and decoding settings. The runs differ only in whether the assigned recovery operation is applied. A task is \emph{prefix-feasible} if it reaches the chosen intervention point and contains enough earlier history to construct the required stale observation. We also verify that the two runs have identical histories before they branch.

Each cohort serves a distinct role. The primary and independent cohorts test whether recovery effects replicate. The mechanism cohort isolates the components of refresh. The evaluation cohort tests CIR.
Some analyses focus on tasks whose factual trajectory succeeds. To check that the conclusions do not depend on this filter, we also report complete cohorts that retain every prefix-feasible task. This gives 106 tasks in the primary cohort and 100 separately executed tasks in the independent cohort. These cohorts cover the available prefix-feasible tasks, not the full ALFWorld distribution. The labels \emph{factual success} and \emph{factual failure} describe the unmodified run. Rescue and harm always compare recovery with no recovery under the same observation condition.

The primary factual-success analysis retains 93 of the 106 \texttt{valid\_unseen} tasks, while the independent cohort contributes 89 factual-success tasks. Mechanism controls use a separate 74-task set, of which 67 pass the paired-analysis protocol checks detailed in Appendix~\ref{app:mechanism_overlap}.

CIR is fitted on the 93 primary tasks and evaluated on a separate set of 76 \texttt{valid\_seen} tasks. The names \texttt{valid\_seen} and \texttt{valid\_unseen} refer to ALFWorld splits. No task appears in both fitting and evaluation. Of the evaluation tasks, 75 are prefix-feasible. Each is tested under four observation conditions, giving 300 evaluation episodes. We apply the fitted models and decision thresholds without further tuning. All tasks remain in the denominator at every delay. If a task has already ended, we carry its final outcome forward and do not apply recovery.

\paragraph{Estimation and metrics.} Our main quantity is the paired difference in task success between recovery and no recovery, measured in percentage points (pp). At delay $d$, we also compare this effect between two-step stale and clean observations. We report the paired outcome counts from Section~\ref{sec:rescue_and_harm}. Confidence intervals use a task-level cluster bootstrap. This keeps all conditions from the same task together in each resample~\citep{efron1979bootstrap}. For learned policies, we additionally report how often the policy intervenes and how often it harms a clean trajectory.

\paragraph{CIR fitting and comparison policies.} CIR fits three \(\ell_2\)-regularized logistic models with \(C=0.1\) to estimate the observation-error probability and the two outcome probabilities in Eq.~\eqref{eq:cir_outcome_probabilities}. Inputs contain only information available before a candidate intervention, including action and reasoning repetition, no-op behavior, observation novelty, consistency between actions and observations, task progress, and context-budget statistics.

To keep model fitting separate from threshold selection, we divide the 93 training tasks into five groups. In turn, we fit the models on four groups and predict the held-out group. This gives every training task an out-of-fold (OOF) prediction from models that did not see that task. We choose the decision thresholds from these predictions and then refit the models on all training tasks. All model parameters and thresholds are fixed before test evaluation. We use \(\lambda=2\), \(\epsilon=0.02\), \(d_{\min}=1\), and thresholds \(\alpha=0.876\), \(\beta=0.052\), and \(\gamma=0\). During evaluation, CIR checks $d\in\{1,2,4\}$ in order and refreshes at most once.

We compare CIR with never refreshing and with a sequential anomaly baseline. The anomaly baseline uses the same error detector at each candidate time but does not predict whether refresh will help. Its threshold is selected from the same OOF predictions, objective, and clean-harm limit used for CIR. We also include a version whose training-time intervention rate is matched to CIR. Neither baseline is adjusted on the test set.
\subsection{Recovery Depends on State and Timing}
\label{sec:exp_state_effect}
\begin{figure}[t]
    \centering
    \includegraphics[width=0.7\linewidth]{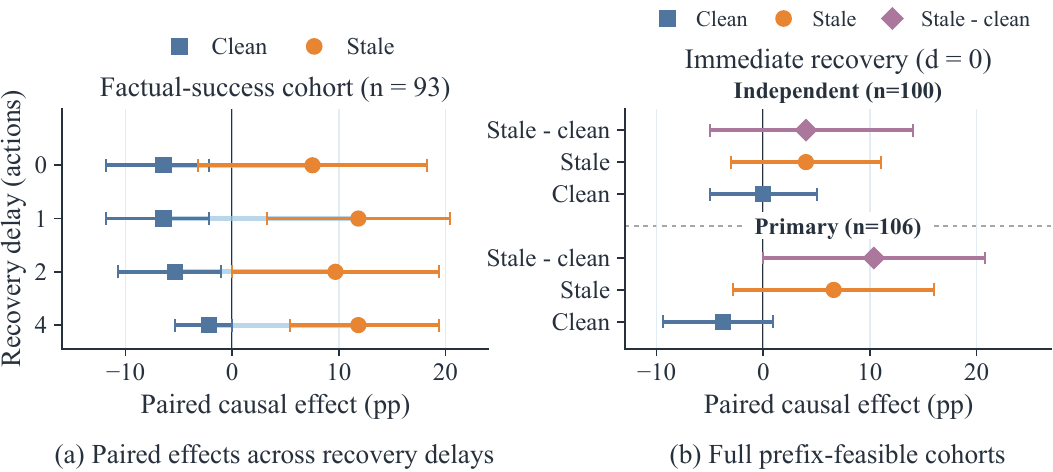}
    \caption{Paired recovery effects in the 93-task factual-success cohort across delays (a) and in the complete prefix-feasible primary (\(n=106\)) and independent (\(n=100\)) cohorts (b). ``Stale'' denotes the two-step stale condition. Error bars are task-level cluster-bootstrap 95\% confidence intervals.}
    \label{fig:causal_effect_heterogeneity}
\end{figure}

We first ask whether the same refresh operation has different effects in different states and at different times. We begin with the 93 primary tasks whose factual trajectory succeeds. Under clean observations, continued execution reproduces a known success, so any failure after refresh is directly interpretable as harm. Figure~\ref{fig:causal_effect_heterogeneity}(a) shows this cohort. An immediate refresh lowers success by \(6.45\) percentage points under clean observations (95\% CI \([-11.83,-2.15]\)). It raises success by \(7.53\) points under two-step stale observations (95\% CI \([-3.23,18.28]\)). The difference between these effects is \(13.98\) points (95\% CI \([2.15,24.73]\)). The stale-minus-clean difference remains positive at later delays. The estimates are \(18.28\), \(15.05\), and \(13.98\) points at delays 1, 2, and 4.

In this cohort, continuing without recovery under a clean observation reproduces the successful factual trajectory, so $Y_i^0(\mathrm{clean})=1$ by construction. Refresh can preserve or harm that success, but it cannot create a rescue. We therefore repeat the analysis with factual failures included.

The independent 89-task factual-success cohort shows the same ordering at every tested delay. The effect under stale observations is more positive than the effect under clean observations, with differences ranging from \(8.99\) to \(14.61\) points. At delay 2, refresh changes success by \(+10.11\) points under stale observations and \(-4.49\) points under clean observations.

We next include every prefix-feasible task in both cohorts. This includes 13 factual failures in the primary cohort and 11 in the independent cohort. Figure~\ref{fig:causal_effect_heterogeneity}(b) summarizes the results. In the 106-task primary cohort, refresh changes success by \(-3.77\) points for clean observations (95\% CI \([-9.43,0.94]\)) and \(+6.60\) points for stale observations (95\% CI \([-2.83,16.04]\)). The difference is \(10.38\) points (95\% CI \([0.00,20.75]\)). In the 100-task independent cohort, the corresponding effects are \(0.00\) and \(+4.00\) points. Their difference is \(4.00\) points (95\% CI \([-5.00,14.00]\)). The estimated effect is therefore more favorable under stale observations in both complete cohorts, including tasks with failed factual trajectories.

The paired counts show what these averages hide. In the complete primary cohort, refresh produces 2 rescues and 6 harms under clean observations, compared with 17 rescues and 10 harms under stale observations. At delay 1 in the factual-success cohort, the positive average under stale observations combines 15 rescues with 4 harms. Under clean observations, refresh turns 6 successes into failures. Appendix~\ref{app:additional_results} gives the full breakdown.

Taken together, these results show that recovery value is not a fixed property of the refresh operation. It depends on the information available to the agent, the state reached after the error, and the time at which recovery is attempted. This explains why a single unconditional rule can hide useful interventions and unnecessary ones behind the same average success rate.

\subsection{Why Does Refresh Help?}
\label{sec:exp_mechanism}

\begin{figure}[t]
    \centering
    \includegraphics[width=0.75\linewidth]
    {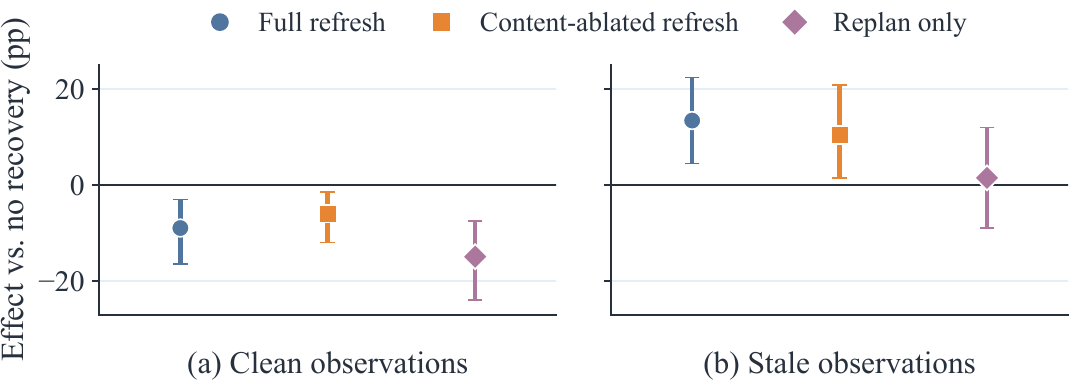}
    \caption{Mechanism controls under clean observations at delay 0 (a) and two-step stale observations at delay 4 (b). Points are effects relative to no recovery. Error bars are task-level cluster-bootstrap 95\% confidence intervals.}
    \label{fig:mechanism_controls}
\end{figure}

We next ask what part of refresh produces the gain. A full refresh queries the environment and shows the returned observation to the agent. A second variant, called content-ablated refresh, makes the same query but hides the returned content. A third variant asks the agent to replan without querying the environment. Under clean observations at delay 0, these operations change success by \(-8.96\), \(-5.97\), and \(-14.93\) percentage points, respectively (Figure~\ref{fig:mechanism_controls}). Under two-step stale observations at delay 4, the changes are \(+13.43\) points (95\% CI \([4.48,22.39]\)), \(+10.45\) points (95\% CI \([1.49,20.90]\)), and \(+1.49\) points (95\% CI \([-8.96,11.98]\)). Full refresh rescues ten tasks. Content-ablated refresh also rescues eight of them. The returned observation is therefore not required for those shared rescues. Appendix~\ref{app:mechanism_overlap} gives the task-level comparison.
This result matters because refresh is often treated as if it were equivalent to supplying better observation content. Our controls show that refresh has two distinct components. It queries the environment and then shows the returned content to the agent. The returned content is unnecessary for most shared rescues. Evaluating only the full operation would miss this distinction.
\subsection{CIR Learns When to Recover}
\label{sec:exp_cir}

We evaluate CIR on all 75 held-out prefix-feasible tasks, whether or not their factual trajectory succeeds. Across 300 episodes, never refreshing succeeds on \(70.33\%\), while CIR reaches \(73.33\%\). The gain is \(3.00\) percentage points (95\% CI \([0.67,5.67]\)), with 11 rescues and 2 harms.

\begin{figure}[t]
    \centering
    \includegraphics[width=0.8\linewidth]{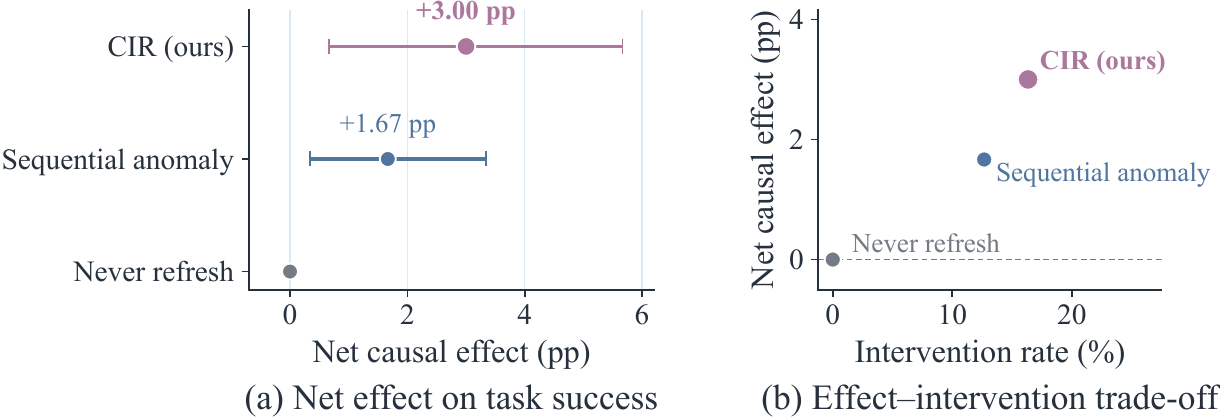}
    \caption{Selective recovery on 75 held-out prefix-feasible tasks. Each task is evaluated under four observation conditions, giving 300 episodes. Points show effects relative to never refreshing. Error bars are task-level cluster-bootstrap 95\% confidence intervals. CIR intervenes in 16.3\% of episodes.}
    \label{fig:cir_policy}
\end{figure}

\begin{table}[t]
    \centering
    \small
    \setlength{\tabcolsep}{5pt}
    \begin{tabular}{lcccc}
        \toprule
        Observation condition & Never & CIR & Effect (pp, 95\% CI) & Rescue / harm \\
        \midrule
        Clean & 81.33\% & 81.33\% & $0.00$ $[0.00,0.00]$ & 0 / 0 \\
        One-step stale & 68.00\% & 69.33\% & $+1.33$ $[0.00,4.00]$ & 1 / 0 \\
        Two-step stale & 60.00\% & 69.33\% & $+9.33$ $[1.33,17.33]$ & 9 / 2 \\
        Missing & 72.00\% & 73.33\% & $+1.33$ $[0.00,4.00]$ & 1 / 0 \\
        \bottomrule
    \end{tabular}
    \caption{Test performance of CIR by observation condition on all 75 held-out prefix-feasible tasks.}
    \label{tab:cir_conditions}
\end{table}

The effect varies by observation condition (Figure~\ref{fig:cir_policy} and Table~\ref{tab:cir_conditions}). CIR never intervenes when the observation is clean, so success is unchanged. Its largest gain occurs when the observation is two steps out of date. Success rises from \(60.00\%\) to \(69.33\%\), an improvement of \(9.33\) points (95\% CI \([1.33,17.33]\)) from 9 rescues and 2 harms. The gains for one-step stale and missing observations are each \(1.33\) points. CIR intervenes in 31 of the 75 two-step stale episodes and in only 1 of the 75 missing-observation episodes. It concentrates recovery where the observed benefit is largest.

CIR intervenes in \(16.3\%\) of episodes. The sequential anomaly baseline intervenes in \(12.7\%\) and improves success by \(1.67\) points (95\% CI \([0.33,3.33]\)), with 6 rescues and 1 harm. The rate-matched anomaly baseline intervenes in \(15.7\%\) and improves success by \(2.00\) points (95\% CI \([0.67,3.67]\)), with 6 rescues and no harms. CIR has the highest point estimate. Its margins over the two anomaly baselines are \(1.33\) points (95\% CI \([-0.33,3.33]\)) and \(1.00\) point (95\% CI \([-0.67,2.67]\)), respectively. The \(3.00\)-point gain over never refreshing is supported by a task-level sign-flip test (\(p=0.034\)).
The rate-matched comparison helps separate selective recovery from simply intervening more often. CIR and the rate-matched anomaly policy refresh a similar fraction of episodes. CIR produces 11 rescues rather than 6 and attains the highest observed success rate. This pattern is consistent with CIR's design. The policy considers the predicted outcome of the intervention, not only whether the current observation appears unusual.
CIR does not trigger on any of the 75 clean episodes, preserving the never-refresh success rate throughout the evaluated clean cohort.

\paragraph{Discussion.} Recovery should be evaluated as a decision, not only as an operation. Paired continuations reveal which trajectories recovery rescues or harms and how its value changes with state and timing. They provide more information than post-recovery success alone. They also show why anomaly detection is insufficient. An unusual observation does not imply that refresh will help. CIR predicts outcomes under refresh and continued execution before acting. A harness can therefore estimate recovery value first and intervene only when the expected value is positive.
\section{Conclusion}
\label{sec:conclusion}

Average success alone hides how recovery changes individual trajectories. Our paired evaluation shows that the value of refresh depends on the observation, the timing, and the trajectory. It also separates the tasks that recovery rescues from those it harms. Controlled refresh variants show that the newly returned observation is not the only source of benefit. CIR turns this evidence into a practical decision. Without retraining the underlying agent, it improves success by \(3.00\) percentage points across 300 episodes from 75 held-out tasks. Its largest gain occurs when observations are more severely outdated, and it attains the highest observed success rate among the policies we compare. These results show how paired causal evidence can support more selective recovery in the harness of LLM agents.

\subsection*{AI use statement}

In this work, we used generative AI tools only for language polishing, including improving grammar, clarity, and fluency of the manuscript. We have not used generative AI tools for generating research ideas, developing the methodology, designing experiments, analyzing results, or drawing scientific conclusions; other disclosure categories are not applicable to this work. We reviewed all AI-assisted edits and verified that they did not alter the technical content or scientific claims. We take full responsibility for the final content of this work, including all text, claims, and artifacts.

\subsection*{Ethics Statement}
All experiments are conducted in controlled benchmark environments and do not involve human subjects or private user data. We do not identify additional ethical risks beyond those commonly associated with LLM agent research.

\bibliography{iclr2027_conference}
\bibliographystyle{iclr2027_conference}

\clearpage
\appendix
\section{Task-Level Overlap between Recovery Operations}
\label{app:mechanism_overlap}

\begin{figure}[t]
    \centering
    \includegraphics[width=0.6\linewidth]{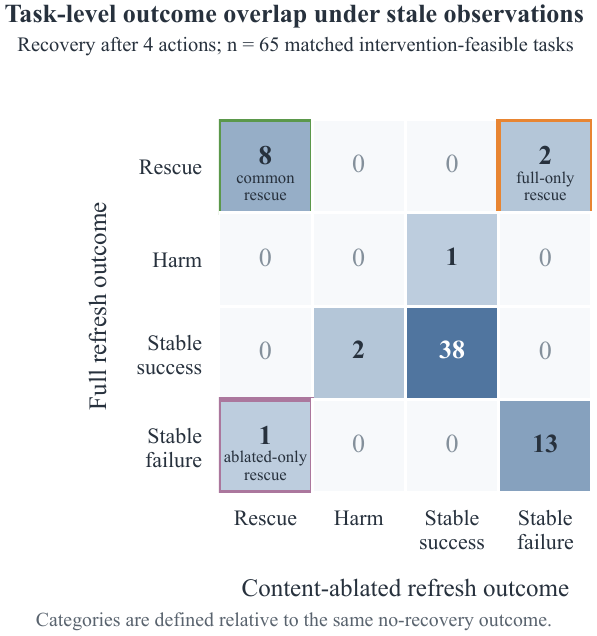}
    \caption{Joint recovery outcomes under two-step stale observations at delay $d=4$. Rows denote full refresh and columns denote content-ablated refresh, each classified relative to the same no-recovery outcome. The analysis includes 65 tasks with matching execution prefixes and excludes two tasks that terminate successfully before recovery. Eight rescues are shared, two occur only under full refresh, and one occurs only under content-ablated refresh.}
    \label{fig:mechanism_outcome_overlap}
\end{figure}

We analyze the joint outcomes of full refresh and content-ablated refresh under two-step stale observations at delay $d=4$. Both outcomes are classified relative to the same no-recovery outcome. The mechanism-control records contain 74 tasks, and 67 satisfy the protocol checks. Two of these tasks terminate successfully before the scheduled recovery, so neither refresh operation is executed. They remain in the 67-task aggregate analysis in Section~\ref{sec:exp_mechanism}, but they are excluded from the joint analysis of executed interventions. Figure~\ref{fig:mechanism_outcome_overlap} shows the joint task-level outcomes. Table~\ref{tab:mechanism_overlap_effects} reports the paired effects. The remaining 65 tasks share the same task identity and exact token prefix before recovery in both branches. No environment-replay mismatches were recorded.

The outcome categories agree on 59 of 65 tasks. These include eight shared rescues, 38 stable successes, and 13 stable failures. The six disagreements include two full-only rescues, one content-ablated-only rescue, two harms avoided by full refresh, and one harm introduced by full refresh relative to content-ablated refresh. The high overall agreement therefore includes 51 tasks whose outcomes neither operation changes. The eight shared rescues show that the returned content is not necessary for these observed benefits.

\begin{table}[t]
    \centering
    \caption{Paired success effects on the 65 tasks where both recovery operations can be executed. Intervals use 10,000 task-level cluster-bootstrap resamples. All values are in percentage points.}
    \label{tab:mechanism_overlap_effects}
    \begin{tabular}{lrr}
        \hline
        Paired comparison & Effect (pp) & 95\% CI \\
        \hline
        Full refresh $-$ no recovery & $+13.85$ & $[4.62,23.08]$ \\
        Content-ablated refresh $-$ no recovery & $+10.77$ & $[1.54,20.00]$ \\
        Full refresh $-$ content-ablated refresh & $+3.08$ & $[-4.62,10.77]$ \\
        \hline
    \end{tabular}
\end{table}

The paired net gains are nine tasks for full refresh versus no recovery, seven for content-ablated refresh versus no recovery, and two for full refresh versus content-ablated refresh. Removing the two already-terminated tasks changes the denominator from 67 to 65 but does not change these gains. This explains the difference from the aggregate effects in Section~\ref{sec:exp_mechanism}. Both refresh variants produce positive gains relative to no recovery. Full refresh has a \(3.08\)-pp point estimate over content-ablated refresh.

\FloatBarrier
\section{Additional Recovery Results}
\label{app:additional_results}

\begin{figure}[t]
    \centering
    \includegraphics[width=0.6\linewidth]{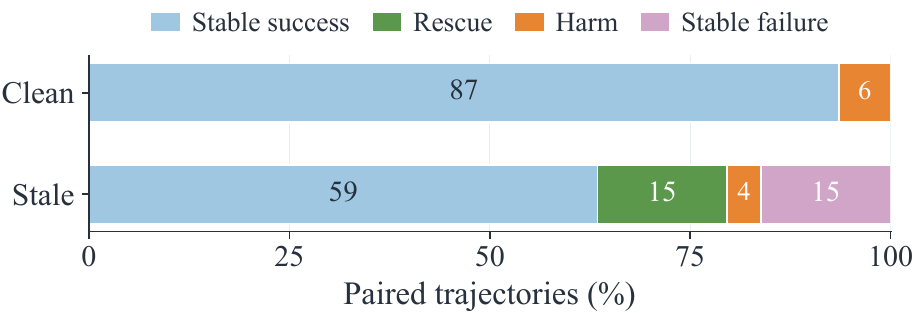}
    \caption{Paired outcomes at recovery delay \(d=1\) in the 93-task factual-success cohort. Each bar partitions trajectories into stable success, rescue, harm, and stable failure.}
    \label{fig:outcome_decomposition}
\end{figure}

Figure~\ref{fig:outcome_decomposition} gives the complete paired-outcome decomposition for the 93-task factual-success cohort at recovery delay \(d=1\). Under stale observations, refresh produces 15 rescues and 4 harms, alongside 59 stable successes and 15 stable failures. Under clean observations, 87 trajectories remain successful and 6 are harmed. This decomposition illustrates how the positive stale-state average arises from trajectory-level changes that aggregate success rates alone conceal.

\end{document}

%% file: math_commands.tex
\usepackage{amsmath,amsfonts,bm}

\def\eqref#1{equation~\ref{#1}}

\def\1{\bm{1}}

\DeclareMathAlphabet{\mathsfit}{\encodingdefault}{\sfdefault}{m}{sl}
\SetMathAlphabet{\mathsfit}{bold}{\encodingdefault}{\sfdefault}{bx}{n}

